\pdfoutput=1

\documentclass[conference]{IEEEtran}

\ifCLASSINFOpdf
\else
\fi

\usepackage{amsmath}
\usepackage{algorithm}
\usepackage{algorithmic}
\usepackage{blindtext, graphicx}
\usepackage{gensymb}
\usepackage{xcolor}
\usepackage{url}
\usepackage{breakurl} 
\usepackage[bookmarks=false]{hyperref}

\usepackage{tikz,lipsum}
\usepackage[labelformat=simple]{subcaption}

\newcommand{\specialcell}[2][c]{%
  \begin{tabular}[#1]{@{}c@{}}#2\end{tabular}}

\DeclareCaptionLabelSeparator{periodspace}{.\quad}
\begin{document}
%
\title{Multi 3D Camera Mapping for Predictive and Reflexive Robot Manipulator Trajectory Estimation}

\author{\IEEEauthorblockN{Justinas Mi\v{s}eikis\IEEEauthorrefmark{1},
Kyrre Glette\IEEEauthorrefmark{1},
Ole Jakob Elle\IEEEauthorrefmark{1}\IEEEauthorrefmark{2} and 
Jim Torresen\IEEEauthorrefmark{1}}
\IEEEauthorblockA{\IEEEauthorrefmark{1}Department of Informatics,
University of Oslo, Oslo, Norway\\
Email: {\{justinm, kyrrehg, oleje, jimtoer\}@ifi.uio.no}}
\IEEEauthorblockA{\IEEEauthorrefmark{2}The Intervention Centre,
Oslo University Hospital, Oslo, Norway}}

\maketitle

\begin{abstract}

With advancing technologies, robotic manipulators and visual environment sensors are becoming cheaper and more widespread. However, robot control can be still a limiting factor for better adaptation of these technologies. Robotic manipulators are performing very well in structured workspaces, but do not adapt well to unexpected changes, like people entering the workspace. We present a method combining 3D Camera based workspace mapping, and a predictive and reflexive robot manipulator trajectory estimation to allow more efficient and safer operation in dynamic workspaces. In experiments on a real UR5 robot our method has proven to provide shorter and smoother trajectories compared to a reactive trajectory planner in the same conditions. Furthermore, the robot has successfully avoided any contact by initialising the reflexive movement even when an obstacle got unexpectedly close to the robot. The main goal of our work is to make the operation more flexible in unstructured dynamic workspaces and not just avoid obstacles, but also adapt when performing collaborative tasks with humans in the near future.

\end{abstract}
\vspace{-0.1cm}
\section{INTRODUCTION}

In many practical applications, industrial robot manipulators are still working "blind" with hard-coded trajectories. This results in the workspace for robots and humans being strictly divided in order to avoid any accidents, which, unfortunately, sometimes still occur. It is often more common to have collision detection systems, which do not always work as expected, rather than collision prevention methods~\cite{ur5collision}. However, \textit{environment-aware robots}~\cite{flacco2012depth}~\cite{rakprayoon2011kinect} are becoming more common, both developed in research and by robot manufacturers themselves (e.g. Baxter by Rethink Robotics)~\cite{fitzgerald2013developing}.

The theme of shared workspace has been researched for many years, however it is still a highly relevant research topic today~\cite{roach1987coordinating}~\cite{leitner2012transferring}. Workspace sharing can be classified as \textit{robot-robot} and \textit{human-robot} systems for task sharing, collaborative or supportive tasks. Normally, in \textit{robot-robot} sharing, controllers of all involved robot systems have direct communication and can coordinate moves easier by knowing the planned trajectories for all the manipulators. We will be focusing on \textit{human-robot} shared workspaces, where sensors are used to observe the environment and adapt the manipulator behaviour according to movements in the workspace, normally caused by human motion.

There have been a number of systems proposed addressing the issue of robot trajectory planning in shared workspaces. One system runs a genetic algorithm using fuzzy logic and defines all obstacles as static while the new trajectory is found~\cite{merchan2006fuzzy}. However, it is not suitable for obstacles moving at higher velocities. Another work provides an analysis of non-verbal cues given by humans and robots, and shows that movement understanding plays an important role in the usability of a system and the \textit{human-robot interaction (HRI)}~\cite{breazeal2005effects}.

Some of the systems for \textit{human-robot} interaction assign the robot arm as a light manipulator, thus, reducing a possible collision force and then using inertia reduction, passivity and parametric path planning~\cite{lew2000interactive}. However, this method leads to light collisions, which, ideally, should be avoided.

With camera systems, especially 3D cameras, becoming more affordable, obstacle detection in the robot workspace becomes easier. Sophisticated methods based on robot manipulator modeling and obstacle motion estimation allow a rapid recalculation of robot trajectories to avoid collision with moving obstacles~\cite{flacco2012depth}. Another system uses multiple Kinect cameras observing the same workspace from different viewing points to avoid collisions, but no definite planning approach was proposed~\cite{lenz2012fusing}. Furthermore, a system was proposed using the historical data of obstacle positions in the workspace of the robot. It avoids the areas which are commonly occluded by a human and plan movement trajectories around them~\cite{hayne2016considering}. However, there is a high risk of a collision if the obstacle appears right in front of the manipulator and is not modeled yet. Such situations can occur, when the obstacle was not seen by the camera before it got too close to the robot, for example due to an occlusion or a blind spot of the camera.


Most of the presented approaches rely on one method, commonly a reactive trajectory re-planning to an unexpected obstacle. We propose a method combining a two layered trajectory planner for a manipulator working in a shared \textit{human-robot} workspace. Multiple 3D cameras are used to observe the workspace from different viewpoints to reduce the chance of occlusions. At the same time, a danger map of the workspace is created indicating the areas which are commonly entered by an obstacle, e.g. person's arm. The system contains two behaviour models. \textit{Reflexive behaviour} immediately reacts to unexpected obstacles appearing close to the robot. \textit{Predictive behaviour} uses the danger map information to predict the probability of an obstacle entering areas of the manipulator workspace and avoids it in advance. The most optimum trajectories are estimated considering the probability of a collision with the obstacle as well as the distance traveled by the end effector of the robot. Collision prevention is done not just for the end effector, but for the whole body of the robot.

One of the possible applications of our proposed method is a surgery assistive robot in an operating theater, where simple tasks like holding a probe or handling surgical tools will be automated. This requires a guaranteed safety with no unexpected impact with a patient or staff around, as well as surrounding equipment. In some cases, there might be multiple robots working in collaboration, for example a robot with a C-arm mounted fluoroscopy scanner working in parallel. Direct communication and motion planning are not always possible, so our proposed method provides an appropriate alternative solution.

Furthermore, with classification of obstacle types, \textit{reflexive behaviour} model can be adapted for collaborative tasks, where a person can hand over objects to the robot or use it for support.


This paper is organized as follows. We present the system setup in Section~\ref{sec:system_setup}. Then, we explain the proposed method in Section~\ref{sec:method}. We provide experimental results in Section~\ref{sec:experiments}, followed by relevant conclusions and future work in Section~\ref{sec:conclusion}.

\section{SYSTEM SETUP}
\label{sec:system_setup}

\begin{figure}[h]
\centering
\includegraphics[width=0.30\textwidth]{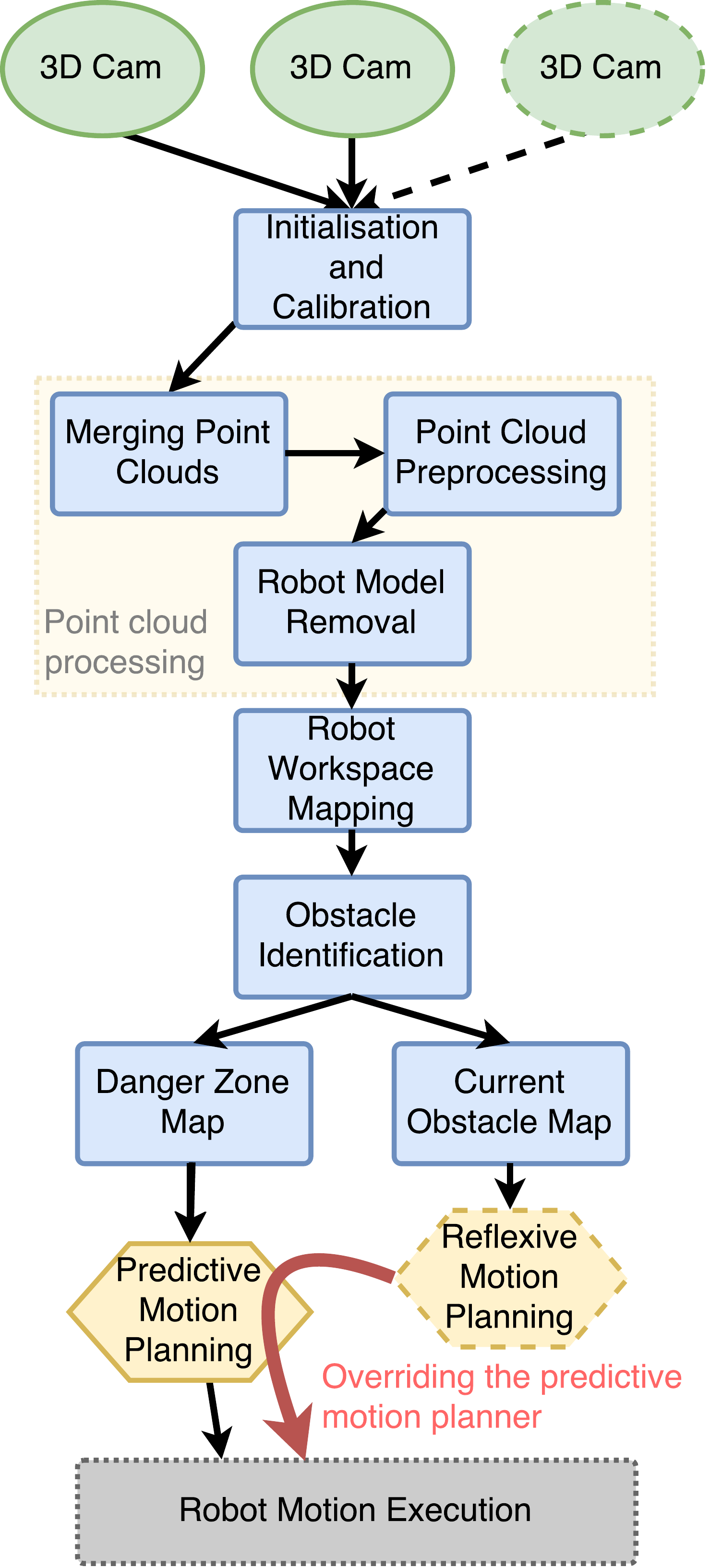}
\caption{Overview of our proposed method. Green ovals represent sensing part, blue rectangles - processing part, yellow hexagons - motion planning and gray rectangle (dashed borders) - motion execution. Reflexive motion planning marked as yellow hexagon (dashed borders) overrides the predictive motion planning when an unexpected obstacle gets close to the robot. There can be a variable number of 3D cameras included in the system.}
\label{fig:system_overview}
\vspace{-0.2cm}
\end{figure}

\subsection{Hardware}

The robotic manipulator being used is UR5 from Universal Robots with \textit{6 degrees of freedom}, a working radius of \textit{850 mm} and a maximum payload of \textit{5 kg}. The repeatability of the robot movements is \textit{0.1 mm}.


In our research we include a low-cost Kinect V2 sensor~\cite{Fankhauser2015KinectV2ForMobileRobotNavigation}. It has been shown to achieve a significantly higher accuracy compared to its predecessor Kinect V1~\cite{amon2014evaluation}. Kinect V2 uses a \textit{time-of-flight (ToF)} approach, using a different modulation frequency for each camera, thus, allowing multiple ToF cameras to observe the same object without any interference~\cite{foix2011lock}. For short-range sensing, an Intel F200 3D camera was mounted on the end-effector to detect any obstacles, which are in close proximity of the end effector~\cite{f200}. Also one Kinect V1 sensor is included in our setup. 
%
In general, any 3D camera, with ROS support, can be used with the system.


\begin{figure}[h]
\centering
\includegraphics[width=0.28\textwidth]{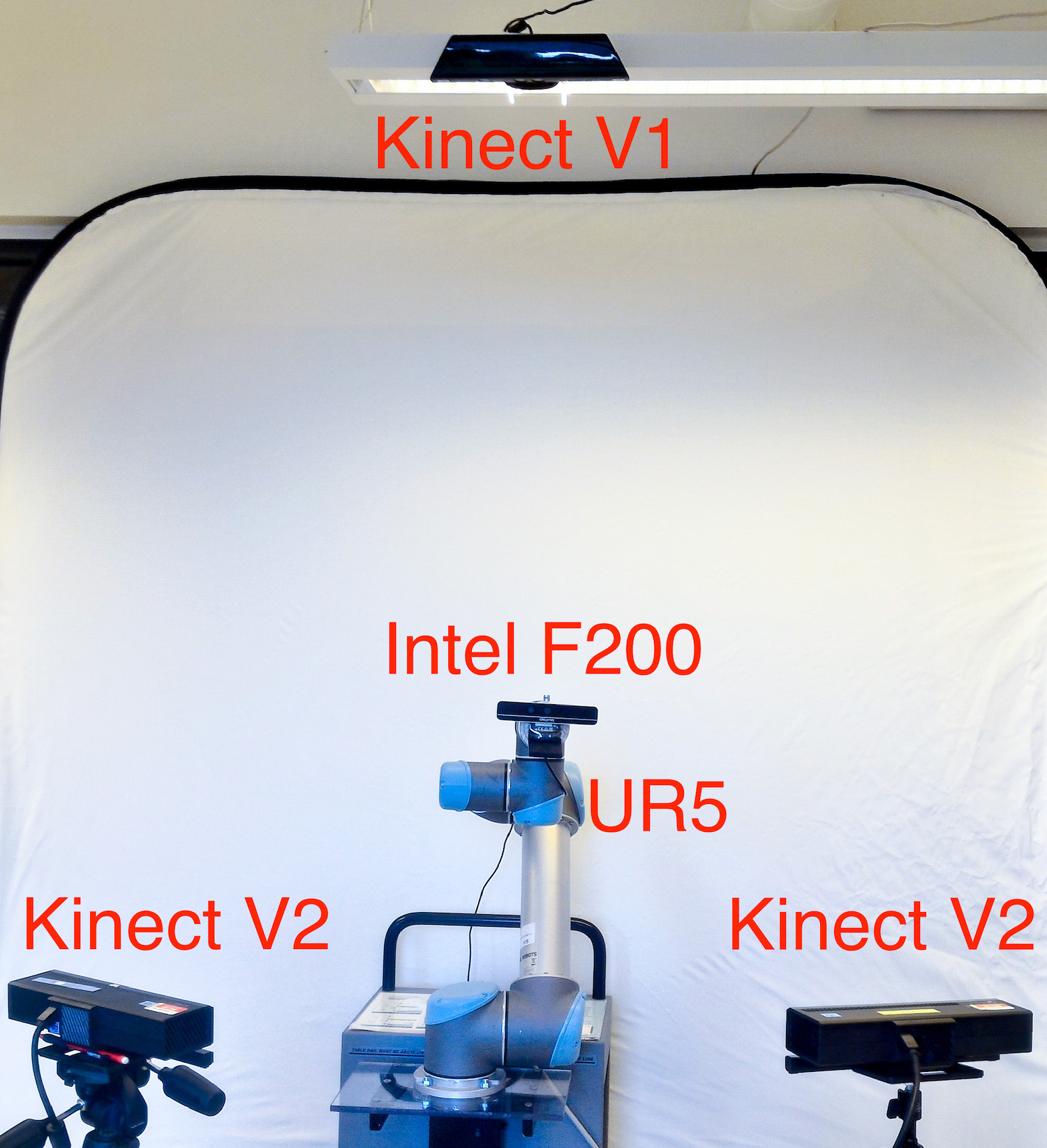}
\caption{Our system setup with overhead Kinect V1, two Kinect V2 cameras placed at different angles observing the front of the robot and an Intel F200 camera mounted on the end effector of the UR5.}
\label{fig:camera_placement}
\vspace{-0.4cm}
\end{figure}

\subsection{Software}

The system software runs on the Robot Operating System (ROS) running on Ubuntu 14.04~\cite{quigley2009ros}. The main advantage of using ROS is its modular design allowing the algorithm to be divided into separate smaller modules performing separate tasks and sharing the results over the network. The workload in our setup was divided over multiple machines.

Kinect V2 is not officially supported on Ubuntu, however, open-source drivers including a bridge to ROS were found to function well, including the GPU utilisation to improve the processing speed of the large amounts of data produced by the sensors~\cite{iaikinect2}. Well tested OpenNI 2 drivers were used to integrate Kinect V1 and Intel F200 into the system.


\vspace{-0.1cm}
\section{METHOD}
\label{sec:method}

\subsection{System Overview}

Our system contains a number of processes working both in series and in parallel as seen in Fig. \ref{fig:system_overview}. Below we present each part of the algorithm in more detail.

\subsection{Calibrating 3D Cameras to the Robot}

The first step of the system setup is to place the 3D cameras around the robot. The goal is to observe the complete robot workspace and to avoid any occlusions. In our case, two Kinect V2 cameras facing the robot were placed, angled at 45{\degree} relative to the robot base, one Kinect V1 overlooking the system from the top and the Intel F200 camera mounted on the end effector of the robot, as shown in Fig. \ref{fig:camera_placement}. However, many different combinations can be used and selection should be made depending on the application.

With fixed camera positions the Eye-To-Hand calibration can be performed to map all the 3D camera coordinate systems to match the robot base coordinate system~\cite{ma2014hand}~\cite{horaud1995hand}. This can be done automatically by placing a calibration board on the robot's end effector and using the proposed automatic calibration procedure~\cite{calib2016jmiseikis}. It uses the estimated checkerboard position and robot joint encoder information to guide the robot movements and cover the field-of-view (FoV) of each of the cameras as much as possible for an accurate Eye-to-Hand calibration.

\begin{figure}[ht]
\hfill
\centering
\begin{subfigure}[t]{.22\textwidth}
    \centering
    \includegraphics[width=\linewidth]{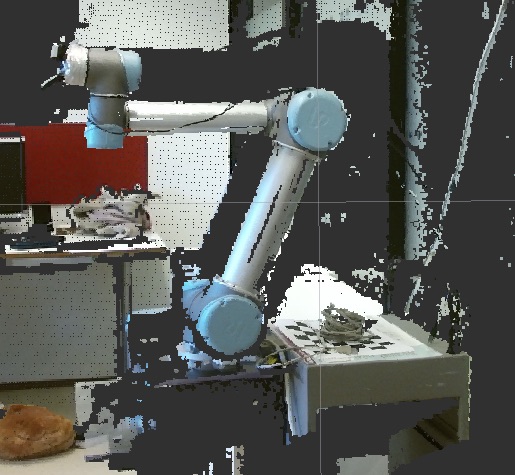}
    \caption{Point cloud}
    \label{fig:pcltooctomap:pcl} 
    \vspace{-0.1cm}
\end{subfigure}
\hfill
\begin{subfigure}[t]{.22\textwidth}
    \centering
    \includegraphics[width=\linewidth]{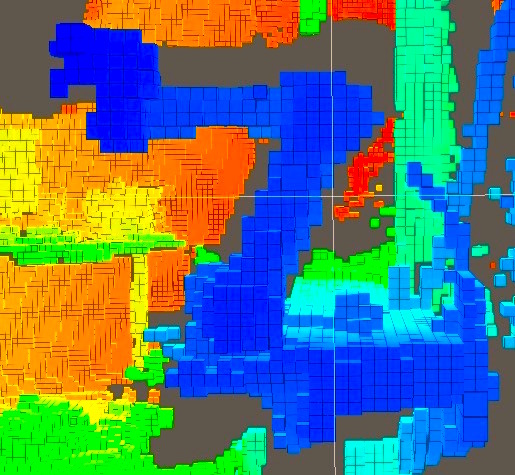}
    \caption{Octomap representing the input point cloud}
    \label{fig:pcltooctomap:octomap}
    \vspace{-0.1cm}
\end{subfigure}

\caption{Octomap created from the Kinect V2 point cloud data of the robot scene. Colorscheme represents the distance of objects from the 3D camera.}
\label{fig:pcltooctomap}
\vspace{-0.2cm}
\end{figure}

The robot automatically performs a number of moves until a precise calibration is achieved. Using the newly calculated transformation matrix describing the positions of the 3D cameras relative to the robot, all the point clouds can be mapped onto a common coordinate frame originating at the robot base.

\subsection{Merging Point Cloud Data}

In order to map the whole workspace of the robot, point clouds from each of the 3D cameras have to be merged. The calibration provides a good estimation of the transformation matrices for accurate merging, but additionally, an Iterative Closest Point (ICP) method is used for fine alignment of all the point clouds~\cite{besl1992method}. The process is performed for each of the cameras. Once the precise transformation matrices have been calculated using the ICP method, they are applied for transformations of all the incoming point clouds. Camera calibration and the ICP method do not need to be repeated unless the cameras or the robot base are moved in relation of each other.


\begin{figure}[h]
\centering
   \begin{subfigure}[b]{0.40\textwidth}
   \includegraphics[width=1\textwidth]{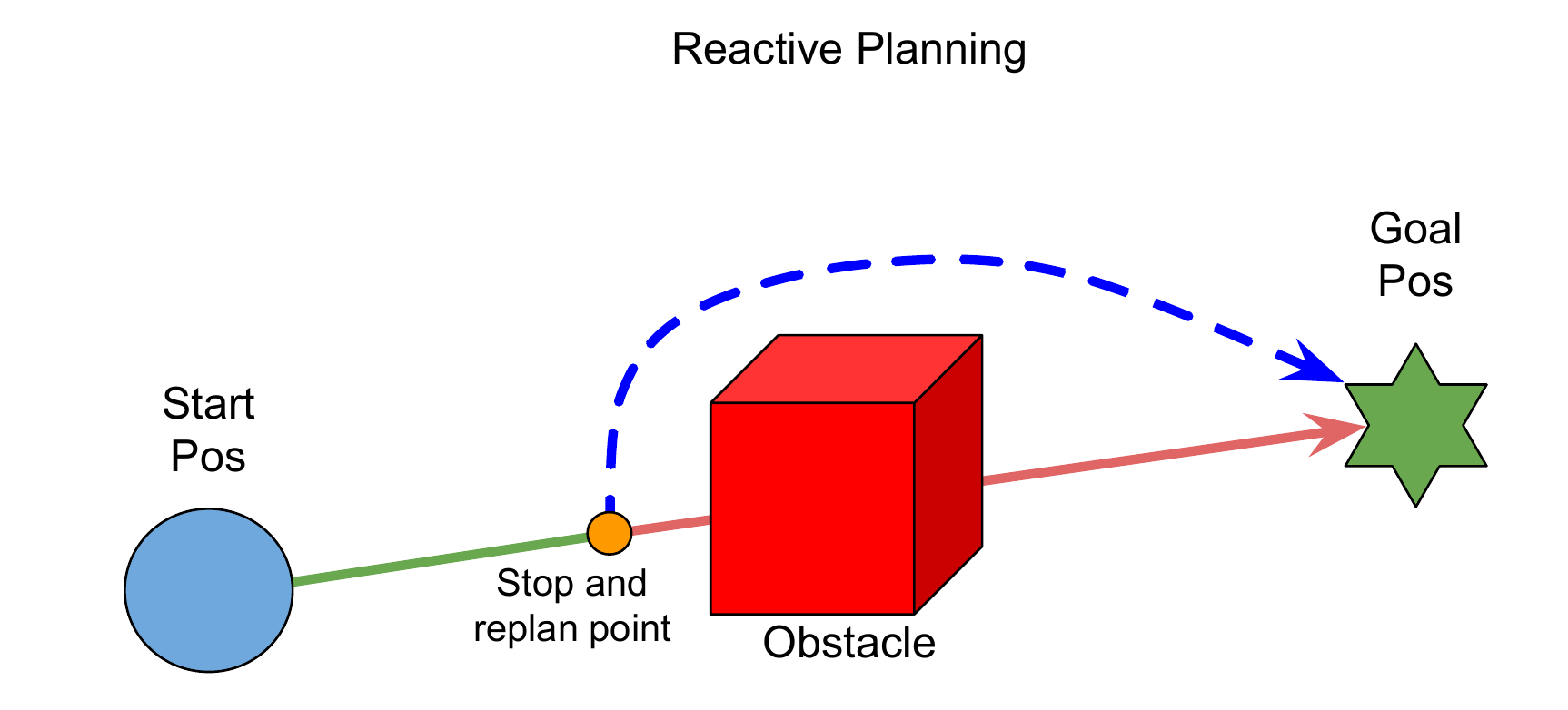}
   \caption{Reactive Behaviour Planning: When the obstacle is present, the robot stops and recalculates the path. Occupied workspace is never crossed.}
   \label{fig:planning_approaches:reactive} 
\vspace{-0.2cm}
\end{subfigure}

\begin{subfigure}[b]{0.40\textwidth}
   \includegraphics[width=1\textwidth]{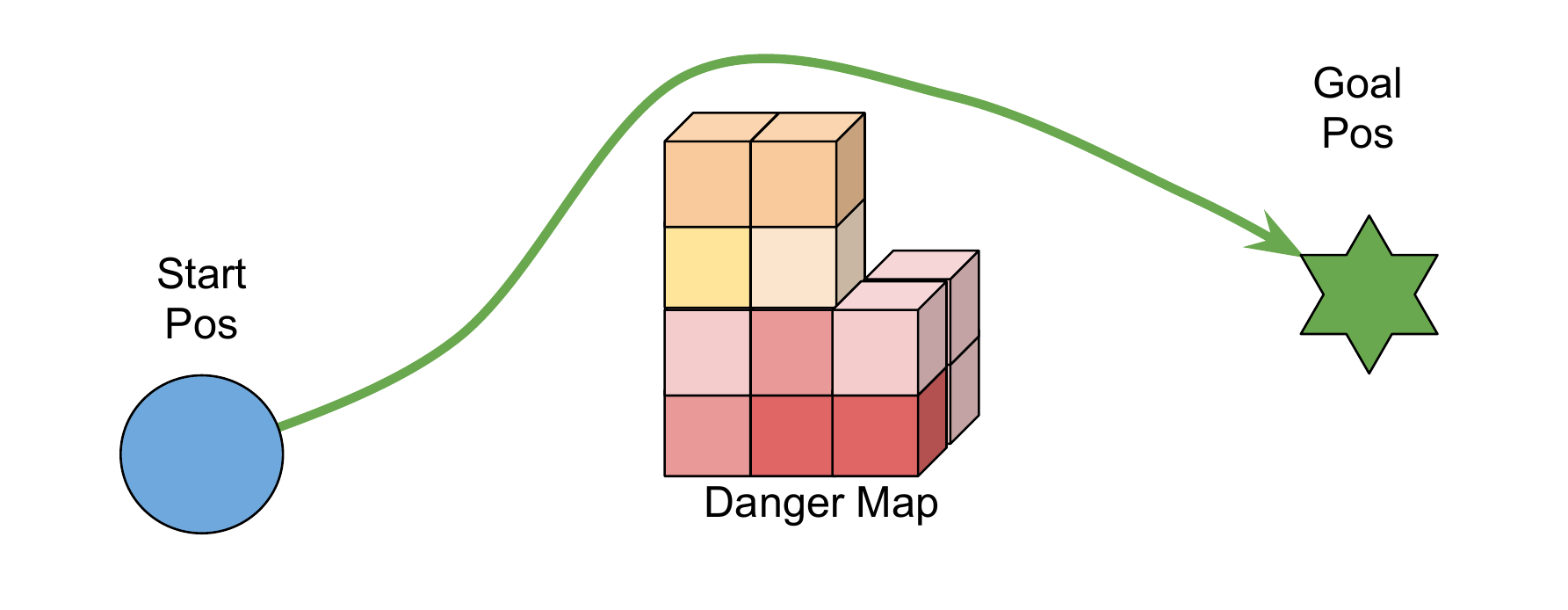}
   \caption{Our method: the trajectory fully avoids, but gets close to the medium risk area in the danger map. If detour is not large, a safe path is chosen over a risky one.}
   \label{fig:planning_approaches:predictive}
\vspace{-0.1cm}
\end{subfigure}

\begin{subfigure}[b]{0.40\textwidth}
   \includegraphics[width=1\textwidth]{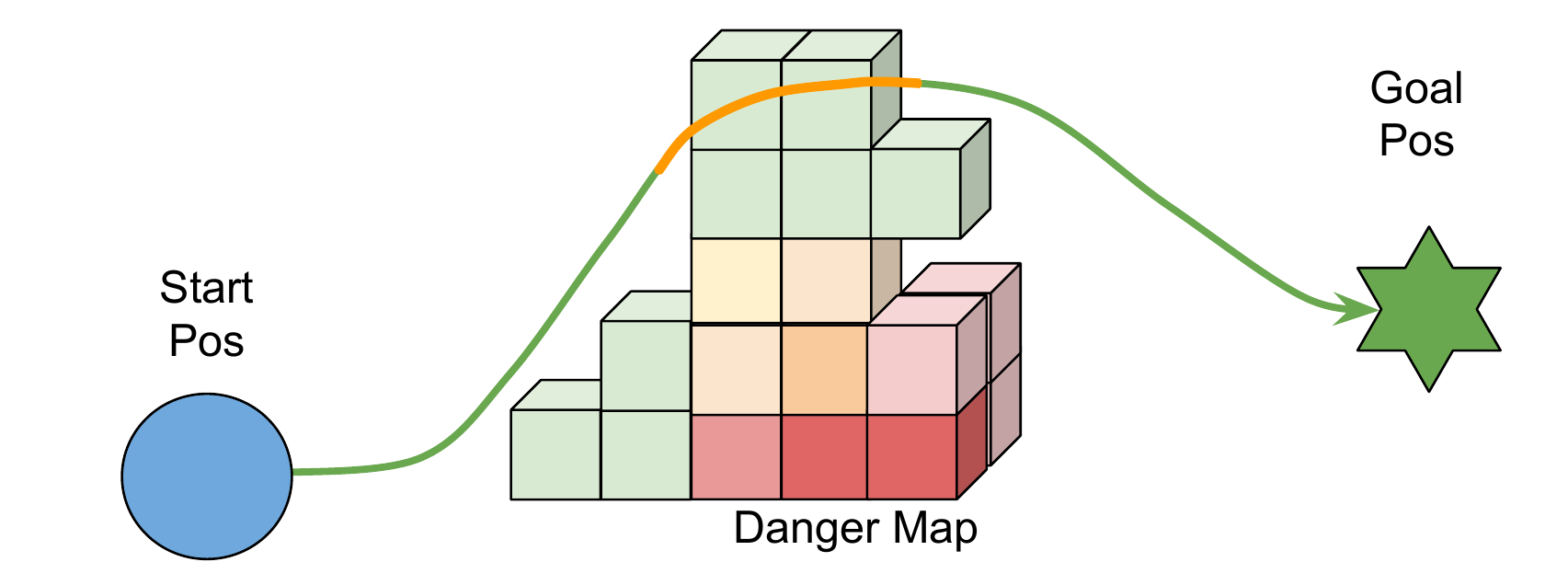}
   \caption{Our method: the trajectory crosses the low risk area in the danger map.}
   \label{fig:planning_approaches:predictivecrossing}
\vspace{-0.2cm}
\end{subfigure}

\begin{subfigure}[b]{0.40\textwidth}
   \includegraphics[width=1\textwidth]{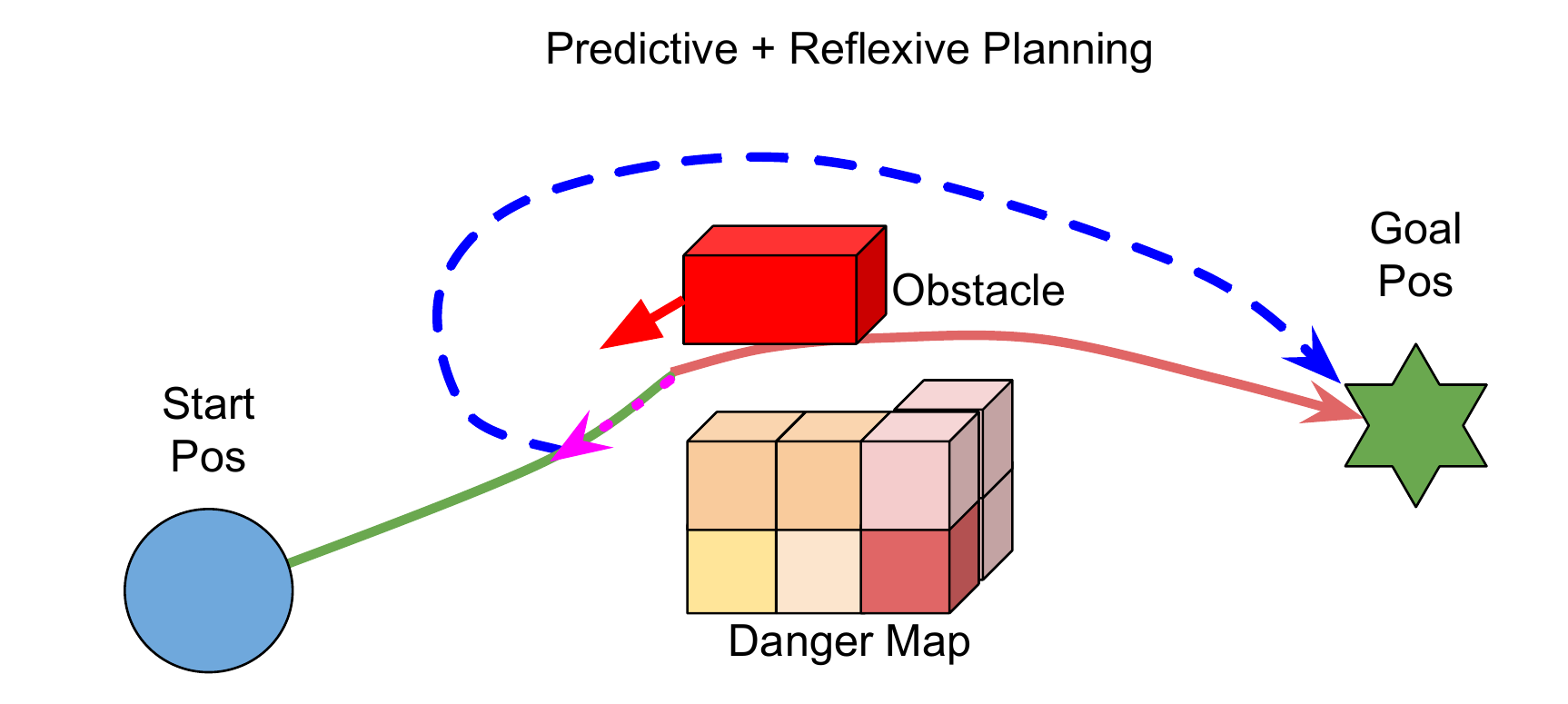}
   \caption{Our method: Predictive and Reflexive Planning. The obstacle gets very close to the robot resulting in the reflexive behaviour being initialised and a new alternative trajectory calculated.}
   \label{fig:planning_approaches:joint}
\end{subfigure}

\caption{Comparison of our proposed joint predictive and reflexive trajectory planning versus a traditional reactive trajectory planning. In danger maps, the scale of risk is from the lowest risk (light green) to the highest risk (red).}
\label{fig:planning_approaches}
\vspace{-0.2cm}
\end{figure}

\subsection{Point Cloud Pre-processing}

In order to increase the processing speed and filter out unwanted noise, a number of pre-processing steps are performed on the input data from each of the 3D cameras.

The first step is to filter out any points in the point cloud data that are too far away from the robot workspace. Knowing that the workspace radius is $850 mm$ from the base of the robot, any points that are further than $1500 mm$ from the robot base are removed as they are not important in our application. This significantly downsizes the point cloud.

Then, any outliers in the point cloud are removed using a Statistical Outlier Removal algorithm~\cite{rusu20113d}. After noisy data has been removed, point clouds can be simplified by down-sampling using a voxel grid filter, which normally reduces the number of points with a minimal loss of information. The voxel grid filter performs a smart down-sampling by sub-diving the space containing point cloud data into a set of volumetric pixels (voxels) and all the points inside each voxel are approximated with the coordinates of their centroid.

\subsection{Removing the Robot Model}

After merging the point clouds, Eye-To-Hand calibrations together with the precise model of the robot arm and its current configuration in space are used to remove the underlying points of the 3D robotic arm model. This step is necessary to avoid false positives on self-collision, as some parts of the robot, seen by the 3D cameras would be interpreted as an obstacle.

It is done by fitting simple shapes, in this case cylinder models on known robot links by taking current angles of all the joint encoders. Cylindrical models are expanded to be $5 mm$ larger than the actual robot links to compensate for any noisy point cloud measurements. Once the model is fitted, any points lying inside the cylinder models are removed under the assumption that the robot itself is represented by these measurements. In order to reduce the computational costs, the shape fitting process is re-done only when the robot moves from the previously fixed position.

\begin{figure}[h]
\centering
\includegraphics[width=0.40\textwidth]{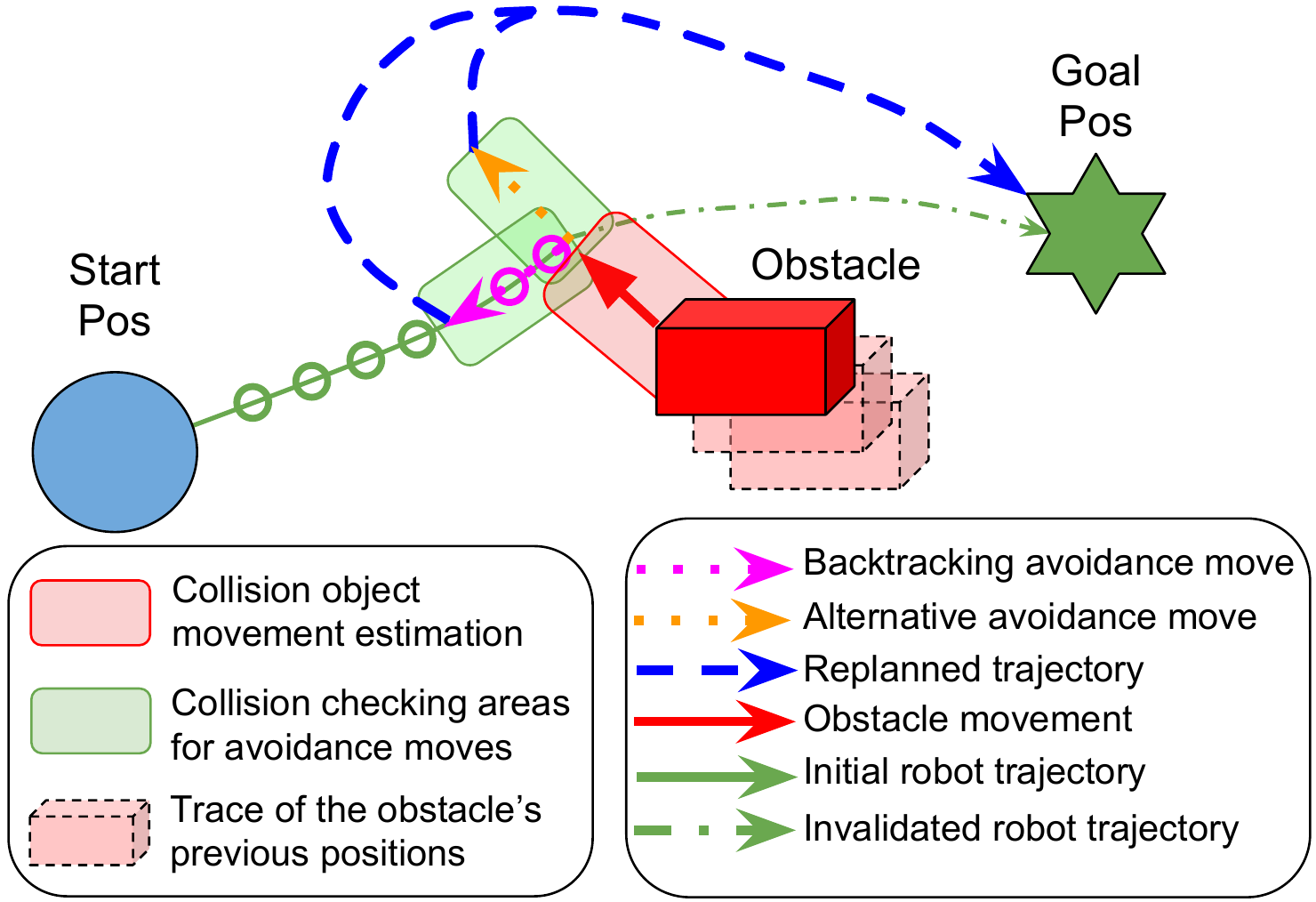}
\caption{Reflexive Behaviour. The planned robot trajectory (green solid line) is blocked by a moving obstacle (marked in red). The first option is to backtrack on the executed trajectory(pink dotted line) until the collision risk is over. If backtracking still results in a collision, the second option is to use the alternative avoidance move and move to the direction opposite from where the obstacle is approaching (dotted orange line). In the meantime, an alternative collision-free trajectory to the goal position is calculated and executed (blue dashed line).}
\label{fig:reflexive_behaviour}
\vspace{-0.4cm}
\end{figure}

\subsection{Mapping}
\label{sub:mapping}

An octomap was chosen as an efficient 3D mapping method of the robot workspace. It is combined of octrees, which are hierarchical data structures for spatial subdivision in 3D. Each node in the octree represents the space contained in a cubic volume, called a voxel. This volume is recursively subdivided into eight sub-volumes until a given minimum voxel size is reached. The minimum voxel size determines the resolution of the octree. The resolution can be dynamically adjusted, both for the whole map, or just parts of it, as each octree branch can be sub-divided into smaller parts~\cite{hornung2013octomap}. This approach enables us to use a simple structure to represent occupied, free and unknown areas in the map. The conversion from the point cloud into an octomap can be seen in Fig.~\ref{fig:pcltooctomap}.

Octomaps can be either binary or full. In binary octomaps, each of their voxels have a binary value, where 1 stands for occupied and 0 for free space. They are suitable for immediate reactions because of the quick processing. While in full octomaps, float values between 1 and 0 are used to describe the probability of voxels being occupied or free, and probabilistic functions can be used to adjust them.

\subsection{Danger Area Identification}
\label{sub:dangerareaidentification}

Any obstacle (e.g. a person entering the workspace of the robot) visible to 3D cameras at the time of the observation is recorded in the octomap as an occupied voxel. As long as the obstacle stays there, the respective voxels will remain occupied. However, when the obstacle is not present anymore in a previously occupied voxel, we do not want to mark it as free immediately. Instead, we introduce a cost function to produce a slow decay, which represents the probability of how risky it is for the robot to enter the area.

\subsubsection{Cost function}

The time dependent cost decay function, shown in Eq.~(\ref{eq:costfunction}), is based on an inverse logarithmic decay to provide a slow decrease at first with an increasing decay the longer the area was not occupied anymore. When the voxel is occupied, its value $C_{voxel,t}$ is reset back to $0.999$.


\begin{equation}
\label{eq:costfunction}
    C_{voxel,t} =  C_{voxel,t-1} + ln(C_{voxel,t-1})*(\Delta t*\alpha)
\end{equation}

Parameter $\alpha$ is used to adjust the decay speed and $\Delta t$ defines the time difference between two calculations, normally determined by the rate of incoming data frames from the camera.
The cost function ensures that the areas in the workspace where obstacles are commonly present and their presence is recurring will be mapped as risky to enter for the robot, and it will attempt to find alternative trajectories through the safe areas to reach the next goal position.

\begin{figure}[h]
\centering
\includegraphics[width=0.35\textwidth]{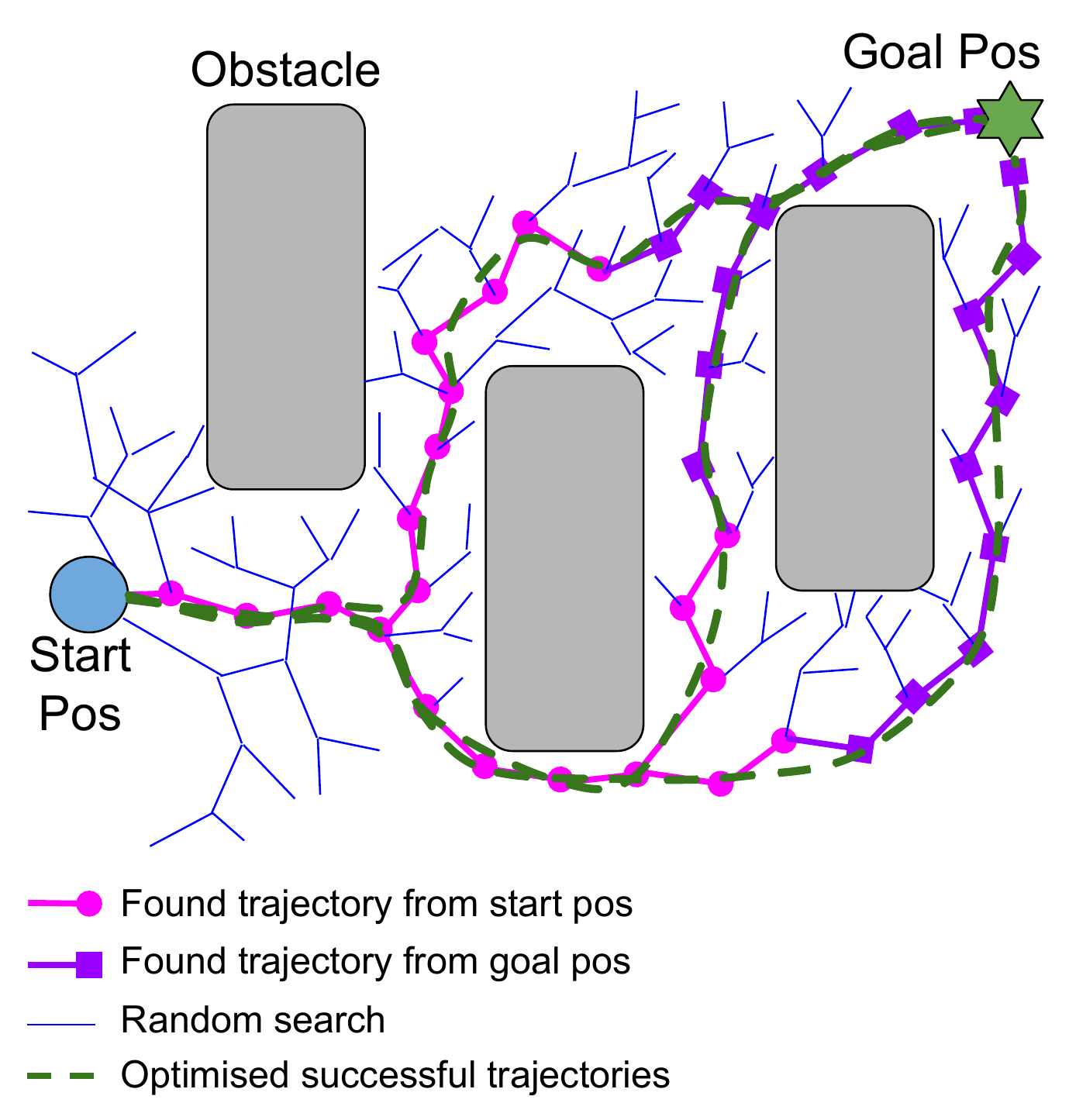}
\caption{Visualisation of a two dimensional RRT-connect trajectory planning algorithm. The method is based on growing Rapidly-exploring Random Trees (marked in thin blue) from the start and goal positions until a connected path is found (marked in pink and purple). Once one or more successful trajectories are found, they are optimised and smoothed (green dashed lines) before they are executed on the robot.}
\label{fig:RRTconnect}
\vspace{-0.2cm}
\end{figure}

\subsection{Robot Motion Planning}

Robot motion planning is based on a two-layered structure: \textit{reflexive} and \textit{predictive behaviour}.

\subsubsection{Reflexive Behavior}

For any immediate danger, a \textit{reflexive behaviour} model is used. Inspired by human behaviour of how we immediately move our hand away from anything that is sharp or burning hot, and only then look at the object and think what to do next. Similarly to this, the robot uses the simplified binary octomap consisting only of currently observed obstacles. If any obstacle is categorised as an immediate danger, the reflexive movement is performed. Last couple of already passed waypoints of the current trajectory are taken as a new goal position, and if the path is free, the movement is immediately executed. Otherwise, if moving back down the previously executed trajectory still results in a collision, an alternative avoidance move is initialised. The movement vector is calculated by taking a vector from the end effector of the robot to the closest point of the obstacle and inverting the direction. The \textit{reflexive behaviour} model is explained in Fig. ~\ref{fig:reflexive_behaviour}. The risk of collision is determined by Eq.~(\ref{eq:riskreflexive}), inspired by the braking distance calculations~\cite{brakingdistance}:

\begin{equation}
\label{eq:riskreflexive}
    D_\textrm{risk}(v, D_\textrm{eucl}) = \frac{|v^2|}{\gamma D_\textrm{eucl}}
\end{equation}

$\gamma$ is a user set parameter, $D_\textrm{eucl}$ is the Euclidean distance between the obstacle and the robot, $v$ is the velocity vector of the obstacle movement. The safe distance is determined by a threshold $T$, which is normally set to 1 and instead the parameter $\gamma$ is adjusted. If $D_\textrm{risk}$ exceeds the threshold $T$, a reflexive motion planning overrides the predictive one.

\begin{equation}
\label{eq:behaviour}
    behaviour = 
\begin{cases}
    reflexive & \text{if } |D_\textrm{risk}|\geq T\\
    predictive & \text{otherwise}
\end{cases}
\end{equation}

\subsubsection{Predictive Behavior}

Independently from the previously described \textit{reflexive behaviour}, a predictive re-planning continuously runs to find the safest and most optimum path to the next goal position. The full octomap, including the calculated danger areas is used for prediction. Each octomap voxel contains the cost (danger) value described in Eq.~(\ref{eq:costfunction}). The motion planner punishes the trajectories, which place any part of the robot in the risky areas (voxels containing non-zero cost). The measure being used for the trajectory evaluation is the accumulated distance through each of the octomap voxels and risk levels added as shown in Eq.~(\ref{eq:trajectorycost}). $C_{traj}$ is the total cost of the trajectory, $D_{traj}$ is the Euclidean distance through the octomap voxel and $C_{voxel}$ is from Eq. (\ref{eq:costfunction}).

\begin{equation}
\label{eq:trajectorycost}
    C_{traj} = \sum_{voxel} (D_{traj} + D_{traj}*C_{voxel})
\end{equation}

This way, a longer trajectory through fully safe areas might be preferred rather than a short and risky one. It has to be noted, that any trajectory execution of the \textit{predictive planner} can be over-ridden by an \textit{reflexive behaviour} planner whenever a high risk obstacle is present. These two planners work in parallel with the reflexive one having higher priority.

\subsubsection{Trajectory Planning and Execution}

\begin{figure}[h]
\centering
\includegraphics[width=0.30\textwidth]{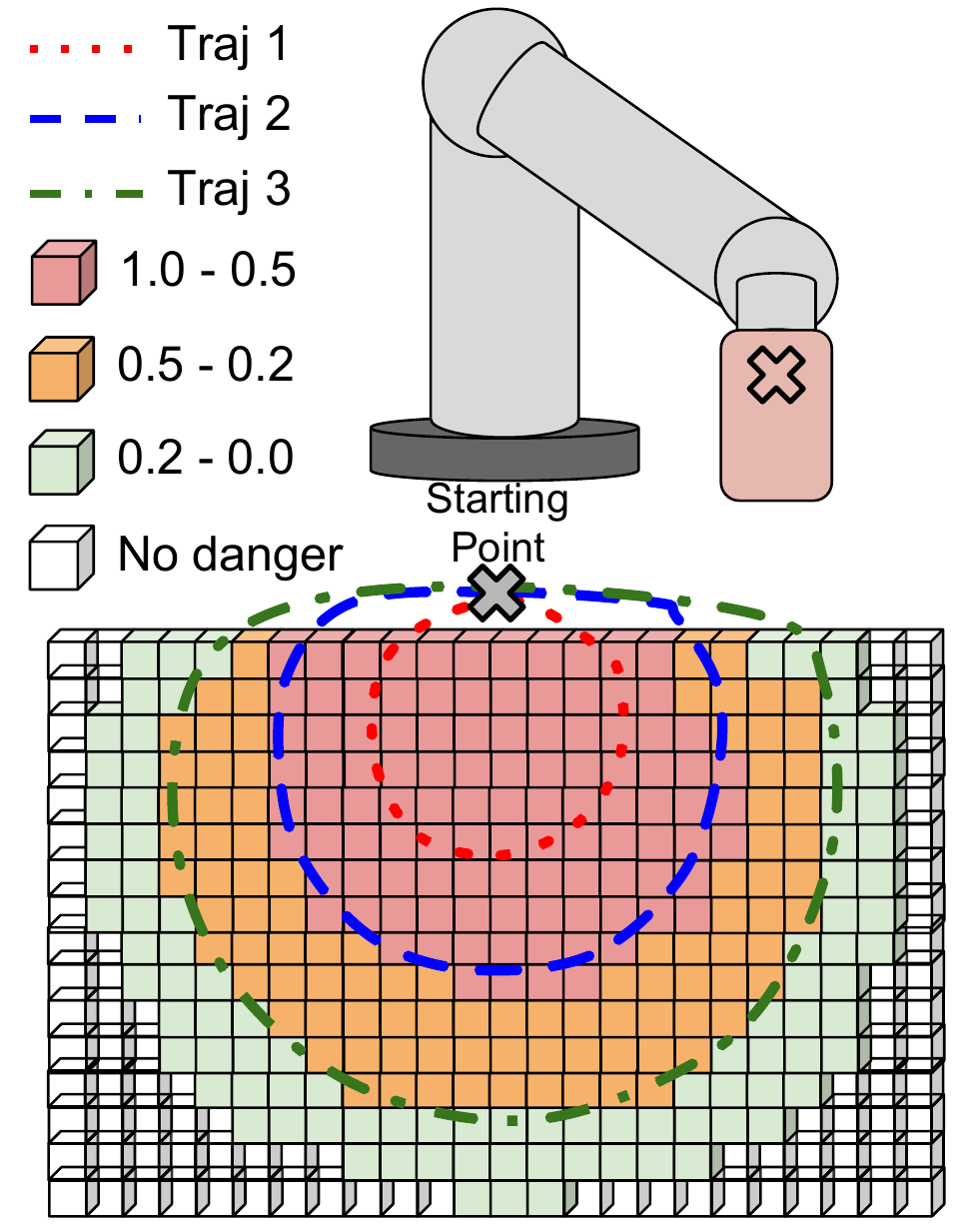}
\caption{Danger map creation test. An object acting as the obstacle was mounted on the end effector of the robot. Then the three indicated trajectories were repeatedly executed in the workspace one after another while the danger map was built up and updated. The average danger cost $C_{voxel}$ of each voxel over the period of the whole experiment is shown with colors red, orange and green indicating high, average and low risk accordingly. Transparent cubes show voxels with zero cost, representing areas without danger.}
\label{fig:mappingtest}
\vspace{-0.2cm}
\end{figure}

\textit{RRT-Connect} motion planner implementation is used for the Cartesian trajectory planning. It is one of the most efficient planners for the UR5 manipulator. \textit{RRT-Connect} stands for Rapidly-exploring Random Trees (RRTs). The method works by incrementally building two Rapidly-exploring Random Trees (RRTs) rooted at the start and the goal configurations. The trees each explore space around them and also advance towards each other through the use of a simple greedy heuristic~\cite{kuffner2000rrt}.

For the trajectory planner to work successfully, the obstacles are precisely modeled in the environment. Then, the free space is defined by calculating all the points, which can be successfully reached given that no part of the robot collides with the obstacles. Then, two the RRTs are initialised both, at the start and goal positions and grown in the free space. The exploration uses the randomly asssigned direction and magnitude of vectors, but they are biased towards the goal position as well as unexplored spaces. If the tree reaches the goal position, or meets the other tree grown from the opposite direction, the successful trajectory has been found. In our case, the exploration is continued until the planning time limit is reached, so more than one successful trajectory can be found. When one or more successful trajectories are found, they are smoothened to avoid choppy robot movements, and a total cost considering the the distance and danger zone crossings (using values from Eq. (\ref{eq:trajectorycost})) is calculated. The most efficient path is executed for the robot to successfully reach the goal position. A two dimensional example of \textit{RRT-Connect} algorithm can be seen in Fig.~\ref{fig:RRTconnect}. 

\begin{figure*}
    \centering
    \begin{subfigure}[t]{0.30\textwidth}
        \includegraphics[width=\textwidth]{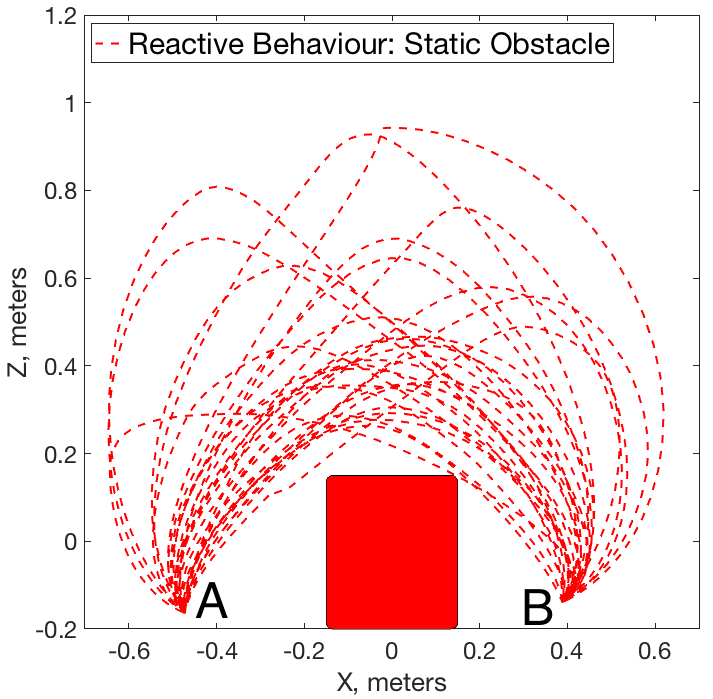}
        \caption{Exp. 1: End-effector position during robot movements between points A and B with reactive behaviour trajectory planning and a static obstacle. Static obstacle is indicated by the red zone.}
        \label{fig:trajectory_comparison:reactstatic}
        \vspace{-0.1cm}
    \end{subfigure}
    ~
    \begin{subfigure}[t]{0.30\textwidth}
        \includegraphics[width=\textwidth]{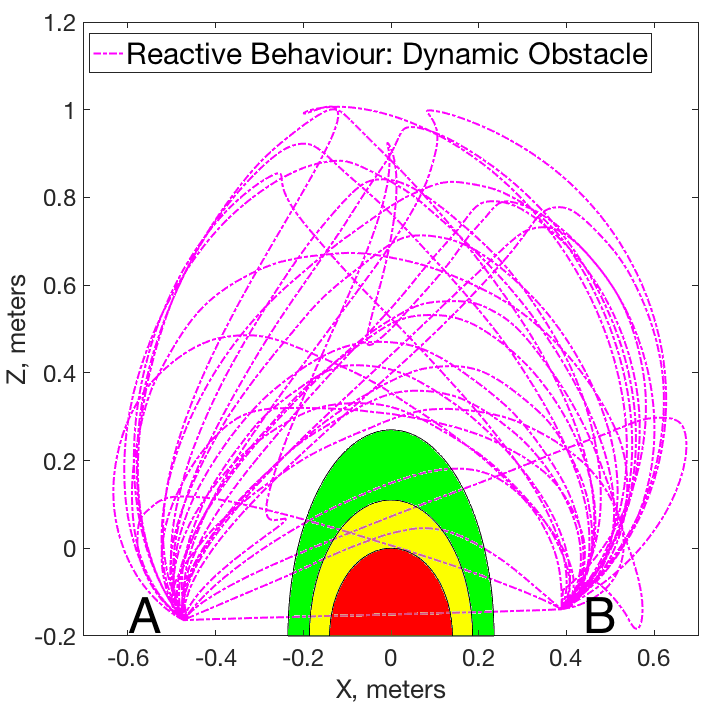}
        \caption{Exp. 2: End-effector position during robot movements between points A and B with a reactive behaviour trajectory planning and a dynamic obstacle. The obstacle was in the areas marked in red, yellow and green with high, medium and low frequency respectively.}
        \label{fig:trajectory_comparison:reactdynamic}
        \vspace{-0.1cm}
    \end{subfigure}
    ~
    \begin{subfigure}[t]{0.30\textwidth}
        \includegraphics[width=\textwidth]{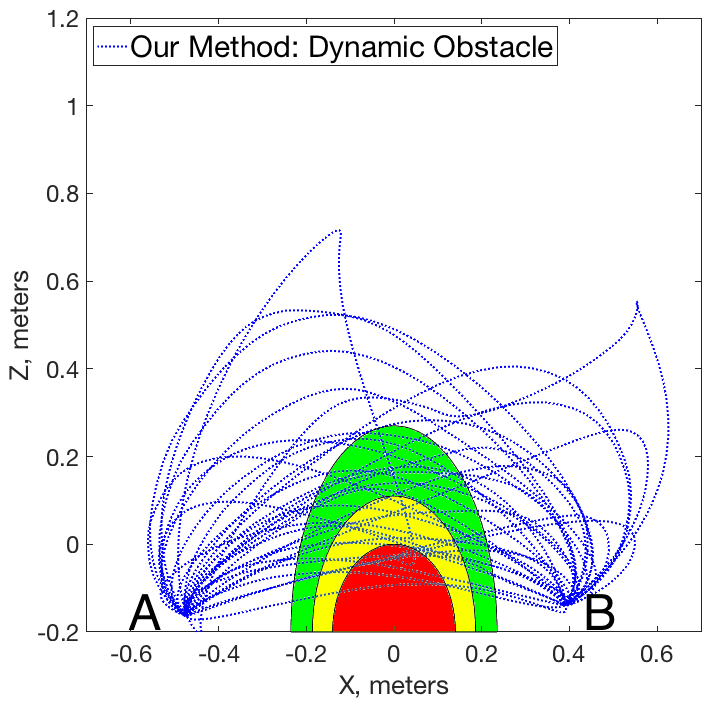}
        \caption{Exp. 3: End-effector position during robot movements between points A and B with our proposed method based on danger maps for trajectory planning and a dynamic obstacle. The obstacle was in the areas marked in red, yellow and green with high, medium and low frequency respectively.}
        \label{fig:trajectory_comparison:our}
        \vspace{-0.1cm}
    \end{subfigure}
    \caption{Comparison of our proposed method and reactive behaviour by tracking the end-effector trajectories under different conditions. The experiment was executed using the presented setup containing one UR5 robot and four 3D cameras. Resulting end-effector trajectories show that our method results in significantly shorter trajectories (compared to a reactive behaviour planning) by taking a calculated risk of crossing part of the danger zone when appropriate instead of taking a long traversal around the areas where a dynamic obstacle might be present. Actual trajectories were planned in 3D, however, for easier visualisation only a front view is shown here, where the difference between the presented methods is the most evident.}
    \label{fig:trajectory_comparison}

\vspace{-0.4cm}
\end{figure*}

This method is suitable, because we can add our calculated cost function to the search space of \textit{RRT-connect} to punish traversals passing through the risk areas calculated in Eq.~(\ref{eq:trajectorycost}).

Motion executions are performed using a new ROS implementation of a velocity based controller with rapid executions and smooth joint accelerations making the robot motions even more human-like. Robot control is done by calculating and directly sending speed commands to each of the robot joints, thus reducing the execution start time to 50-70 ms compared to around 170 ms using the traditional ROS UR5 drivers. The new controller is usable out-of-the-box and compatible with traditional \textit{MoveIt!} trajectory execution~\cite{andersen2015optimizing}.

\begin{figure*}
    \centering
    \begin{subfigure}[t]{0.27\textwidth}
        \includegraphics[width=\textwidth]{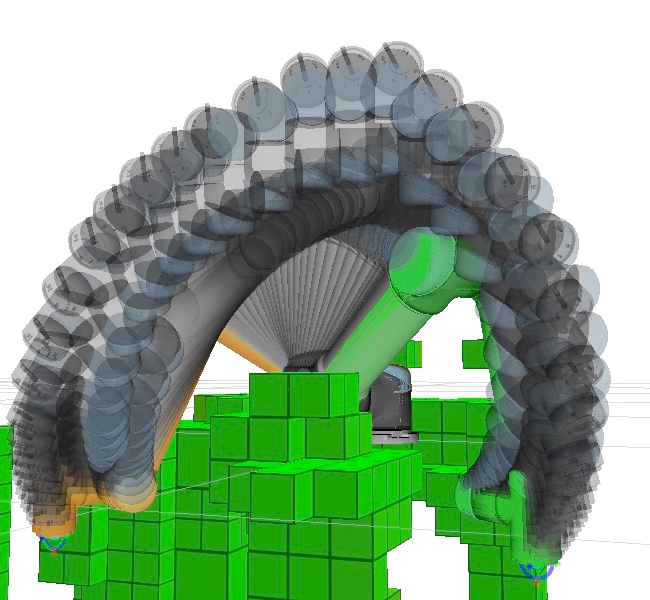}
        \caption{Reactive behaviour: Planned trajectory when the obstacle is determined as static object blocking the direct path.}
        \label{fig:real_robot_trajectories:noobstacle}
    \vspace{-0.1cm}
    \end{subfigure}
    ~
    \begin{subfigure}[t]{0.27\textwidth}
        \includegraphics[width=\textwidth]{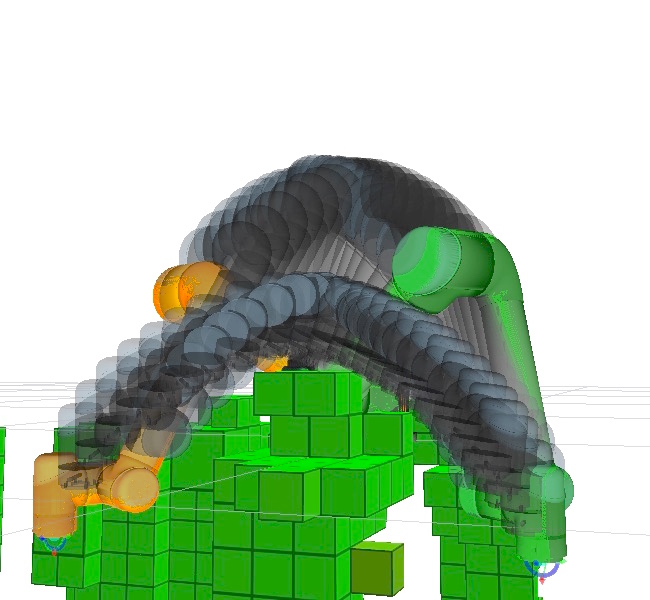}
        \caption{Our method: Planned trajectory while avoiding the danger zone caused by obstacle occasionally entering the workspace.}
        \label{fig:real_robot_trajectories:obstacle}
    \vspace{-0.1cm}
    \end{subfigure}
    ~
    \begin{subfigure}[t]{0.27\textwidth}
        \includegraphics[width=\textwidth]{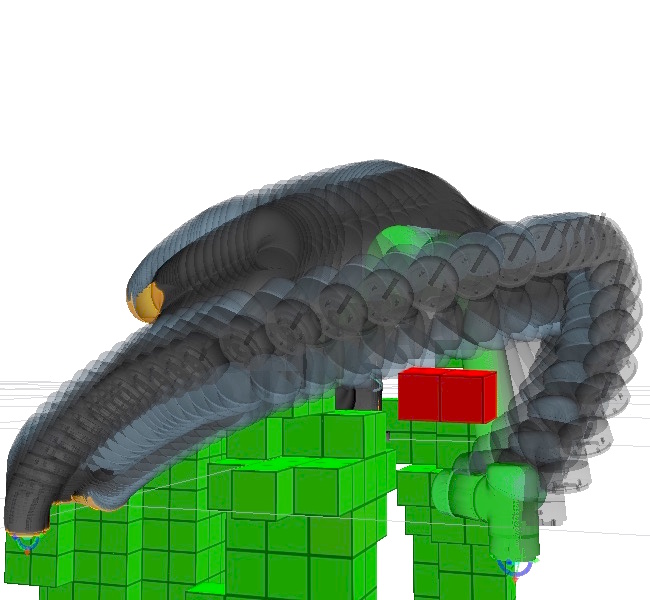}
        \caption{Our method: Planned trajectory while avoiding the danger zone and reflexive behaviour initialised by an obstacle (in red).}
        \label{fig:real_robot_trajectories:predictive}
    \vspace{-0.1cm}
    \end{subfigure}
    \caption{Visualisation of the executed movements on the UR5 robot. Octomaps use the data from 3D cameras and different trajectory planning approaches were compared. Green robot shadow indicates the start position and yellow robot shadow indicates the goal position. Robot shadow trail indicates the trajectory.}
    \label{fig:real_robot_trajectories}
\vspace{-0.4cm}
\end{figure*}

\vspace{-0.1cm}
\section{EXPERIMENTS AND RESULTS}
\label{sec:experiments}

To test our method, we split the experiments into two main parts. First, we evaluate the danger map construction by using a number of repetitive trajectories of an obstacle moving through the workspace. Secondly, the full system is evaluated by using both static and dynamic obstacles in the workspace. All the experiments were conducted using real hardware: the UR5 robot manipulator, two Kinect V2, one Kinect V1 and one Intel F200 3D cameras, as shown in Fig.~\ref{fig:camera_placement}. All the processing was done in real time two computers working in parallel connected with $1000$ Mbit/s internal network.

\subsection{Parameter Values}

Algorithm parameter values were found by \textit{trial-and-error} during the development of the presented method. They have proven to provide good accuracy, while still keeping the processing time low enough for fast and smooth execution of the robot movements. One set of parameter values was used in the current experiments and they are summarised in Table~\ref{table:params}.

\begin{table}[h]
\caption{Parameter values used in our experiments.}
\label{table:params}
\centering
    \begin{tabular}{| l | l |}
    \hline
    
    \textbf{Parameter} & \textbf{Value}\\ \hline
    Cost Function: $\alpha$ & 0.3\\ \hline
    Collision Risk Function: $\gamma$ & 1.5\\ \hline
    Octomap Voxel Size & 0.05 meters \\ \hline
    Backtracking Move Magnitude & 10\% of the executed trajectory \\ \hline
    Max Robot Joint Speed & 50\% of the maximum \\ \hline
    Max Robot Joint Acceleration & 60\% of the maximum \\ \hline
    Kinect V2 Refresh Rate & 15 FPS \\ \hline
    Kinect V1 Refresh Rate & 15 FPS \\ \hline
    Intel F200 Refresh Rate & 30 FPS \\ \hline
    \textit{RRT-Connect}: Max Planning Time & 1 sec \\ \hline
    \textit{RRT-Connect}: \# Planning Attempts & 3 \\ \hline
    \textit{RRT-Connect}: Orientation Tolerance & 0.1 rad \\ \hline
    \textit{RRT-Connect}: Position Tolerance & 0.01 meter \\ \hline
    
    \end{tabular}
\vspace{-0.2cm}
\end{table}

\subsection{Danger Map Construction}

Danger map creation was tested separately by mounting a $10$ cm by $20$ cm object acting as an obstacle on the end effector of the robot and executing pre-defined trajectories in the workspace as shown in Fig.~\ref{fig:mappingtest}. The trajectories were close to a circular shape and executed in a continuous order one after another. The whole process was repeated $20$ times. Trajectory 1 had a radius of $20 cm$, trajectory 2 had a radius of $32 cm$ and trajectory 3 had a radius of $46 cm$. The execution time of each trajectory was $9.1$ sec, $14.69$ sec and $21.06$ sec respectively. With the current cost function parameter $\alpha$ set to $0.3$, the decay time of the voxel from occupied to free is $24.48 sec$.

Throughout the process, the cost values $C_{voxel}$ of all the voxels were tracked and averages calculated. This resulted in a complete danger zone, which could be sub-divided into three areas by the average costs representing the severity of possible collisions. As expected, the outside circle, had the lowest risk with values ranging between $0.0$ and $0.2$, the middle circle had a medium risk with values ranging between $0.2$ and $0.5$ and the inner circle had the highest average risk with values ranging between $0.5$ and $1.0$. The result was as expected, where the voxels falling in the inner area were occluded by robot's body when executing the outer trajectory 3. Results can be seen in Fig.~\ref{fig:mappingtest} with all three trajectories marked and cube colors indicating the different average risk levels.

\subsection{Operation of The Whole System}

Operation of the whole system was tested by planning and executing the trajectory between the two pre-defined points A and B in the workspace. Our proposed method was compared against a simple reactive behaviour based on the same \textit{RRT-Connect} trajectory planner which is used as a baseline. In the first experiment the reactive behaviour planner was used with a static obstacle blocking a direct path between points A and B. The second experiment the reactive behaviour planner was used with a dynamic obstacle randomly moving (moved manually by the operator) into the area blocking a direct path between points A and B. And in the third experiment, our proposed method, based on predictive and reflexive behaviour, was tested by using identical dynamic obstacle randomly moving into the area blocking a direct path between points A and B. In total, $15$ executions of return A-B-A trajectories were executed in each of the experiments.

Because the \textit{RRT-Connect} algorithm is based on random elements and provides different solution every time, the planned trajectories were different for every movement. In the experiment 1, the reactive behaviour planner successfully traversed around the static obstacle while keeping a safe margin between the robot and the obstacle, as seen in Fig.~\ref{fig:trajectory_comparison:reactstatic}.

In the experiment 2, the obstacle randomly entered the indicated workspace with different frequency. Trajectories created by the reactive behaviour planner were significantly more random and spread all around the workspace. It was caused by some attempts of moving more or less directly between the two points and then reacting to a blocked path by the dynamic obstacle. In such cases, the robot stopped and quickly re-planned the trajectory. However, due to very limited planning time, the new trajectory was often not optimum and took a long and unnecessary detour with high safety margins. Longer allowed re-planning times would make new trajectories more optimal, however, the whole execution time would be likely to be even longer. Resulting trajectories of the experiment 2 can be seen in Fig.~\ref{fig:trajectory_comparison:reactdynamic}.

In the experiment 3, the dynamic obstacle was acting in an identical manner as in the experiment 2. Our proposed trajectory planner was constructing a danger map and using it to plan the trajectories between the points A and B, and the planned paths were significantly shorter and smoother compared to the results using the reactive behaviour. Low risk areas were often crossed with occasional crossings of the medium risk areas and a few crossings of the high risk areas. The reflexive behaviour was initialised only in two instances, both times the robot successfully moving away from the obstacle and avoiding the collision. On average, our proposed method provided the smoothest and shortest trajectories compared to the other two experiments. Resulting trajectories of the experiment 3 can be seen in Fig.~\ref{fig:trajectory_comparison:our}.

Another evaluation criteria can be trajectory planning, optimisation and execution time. Our method demonstrated in the experiment 3 has over three times faster average trajectory execution time compared to the reactive behaviour (experiment 2) when the dynamic obstacle was present. The big difference mainly appears due to multiple stops by the reactive behaviour planner to replan the motion when the obstacle blocks the trajectory being executed. Normally it moves in the most direct free trajectory without considering the historical data, which in our method is considered by looking at the danger map. Furthermore, our proposed method is close to $20\%$ quicker than the experiment 1, where the trajectory planning was performed to avoid the static obstacle. No significant differences were observed in planning and trajectory optimisation times between the three experiments, however our proposed method has the shortest time by a small margin. Timing results of all three experiments and a comparison against the obstacle-free \textit{direct trajectory} can be found in Table~\ref{table:timing_results}.

\begin{table}[h]
\caption{Trajectory planning and execution timing results containing averages and standard deviations of the path planning time, the path optimisation (smoothing) time and the path execution time. Results are compared against \textit{direct trajectory}, which had no obstacles present.}
\label{table:timing_results}
\centering
    \begin{tabular}{| l | l | l | l | l |}
    \hline
     & \textbf{Exp. 1} & \textbf{Exp. 2} & \textbf{Exp. 3} & \textbf{Direct} \\ \hline
    \textbf{Planning time (s)} & \specialcell{$0.0183$ \\ $\pm 0.0099$} & \specialcell{$0.0217$ \\ $\pm 0.0095$} & \specialcell{$0.0141$ \\ $\pm 0.0072$} & \specialcell{$0.0145$ \\ $\pm 0.0057$}\\ \hline
    
    \textbf{Optim. time (s)} & \specialcell{$0.0178$ \\ $\pm 0.0102$} & \specialcell{$0.0130$ \\ $\pm 0.0089$} & \specialcell{$0.0129$ \\ $\pm 0.0077$} & \specialcell{$0.0006$ \\ $\pm 0.0002$} \\ \hline
    
    \textbf{Execution time (s)} & \specialcell{$10.6085$ \\ $\pm 4.0766$} & \specialcell{$26.6663$ \\ $\pm 14.7546$} & \specialcell{$8.2952$ \\ $\pm 4.3454$} & \specialcell{$4.994$ \\ $\pm 0.5421$} \\
    \hline
    \end{tabular}
\vspace{-0.2cm}
\end{table}

Some selected example trajectories of each method with the full UR5 body visualised are shown in Fig.~\ref{fig:real_robot_trajectories} for easier visualisation of how the execution looks on the real system.


\vspace{-0.1cm}

\section{CONCLUSIONS AND FUTURE WORK}
\label{sec:conclusion}

In this paper we have presented the predictive and reflexive trajectory planning method for a robot manipulator based on a danger map constructed using a multi 3D camera system. It is designed to function better in the workspaces where unknown dynamic obstacles are present. In the experiments, it has proven to be more effective than the traditional reactive trajectory planner, and still being able to avoid collisions with unexpected obstacles getting close to the robot body. 

Our proposed system contains a combination of many methods working in parallel, each one of them having a set of tunable parameters, affecting the performance of the whole system. In our tests, a trial-and-error method was used to find a good combination of parameter values. However, in the future work, we plan to use AI systems, like evolutionary algorithms, to automatically compute the best parameter value set for case and make the system even more adaptive to changing conditions by learning over time.

Additionally, obstacle classification will allow the robot to react differently depending whether a person is approaching the robot, or some other object. Also, making a difference between bare hand or somebody holding a tool, especially a sharp one, a different size safety zone should be used and the robot should engage in different behaviour.

For collaborative tasks, the contact between the robot and human might be beneficial. With modeling and understanding the behaviour of a person sharing the workspace, joint tasks for object handover or robot working as a support as well as directing certain tools to a required area will become possible.





\vspace{-0.1cm}

\section*{ACKNOWLEDGMENT}
This work is partially supported by The Research Council of Norway as a part of the Engineering Predictability with Embodied Cognition (EPEC) project, under grant agreement 240862

\vspace{-0.1cm}

\bibliographystyle{IEEEtran}
\bibliography{IEEEexample}

\end{document}